\documentclass{article}

\PassOptionsToPackage{numbers}{natbib}
 \usepackage[dblblindworkshop,final]{neurips_2025}
 \usepackage{graphicx}
\workshoptitle{Foundation Models for the Brain and Body}

\usepackage[utf8]{inputenc} 
\usepackage[T1]{fontenc}    
\usepackage{hyperref}       
\usepackage{url}            
\usepackage{booktabs}       
\usepackage{amsfonts}       
\usepackage{nicefrac}       
\usepackage{microtype}      
\usepackage{xcolor}         
\usepackage[letterpaper]{geometry}

\title{Region-Aware Reconstruction Strategy for Pre-training fMRI Foundation Model}

\author{%
Ruthwik Reddy Doodipala$^{1,2^*}$ \quad Pankaj Pandey$^{1^*}$ \quad Carolina Torres Rojas$^1$ \\
\textbf {Manob Jyoti Saikia}$^2$ \quad \textbf{Ranganatha Sitaram}$^1$ \\ 
$^1$St. Jude Children's Research Hospital \quad $^2$The University of Memphis\\
\texttt{\{rdoodipa,ppandey,ctorre51,rsitaram\}@stjude.org} \\
\texttt{msaikia@memphis.edu}
}

\begin{document}
\maketitle
\begingroup
\renewcommand\thefootnote{}\footnotetext{$^*$ Equal contribution.}
\addtocounter{footnote}{-1}
\endgroup

\begin{abstract}
The emergence of foundation models in neuroimaging is driven by the increasing availability of large-scale and heterogeneous brain imaging datasets. Recent advances in self-supervised learning, particularly reconstruction-based objectives, have demonstrated strong potential for pretraining models that generalize effectively across diverse downstream functional MRI (fMRI) tasks. In this study, we explore region-aware reconstruction strategies for a foundation model in resting-state fMRI, moving beyond approaches that rely on random region masking. Specifically, we introduce an ROI-guided masking strategy using the Automated Anatomical Labelling Atlas (AAL3), applied directly to full 4D fMRI volumes to selectively mask semantically coherent brain regions during self-supervised pretraining. Using the ADHD-200 dataset comprising 973 subjects with resting-state fMRI scans, we show that our method achieves a 4.23\% improvement in classification accuracy for distinguishing healthy controls from individuals diagnosed with ADHD, compared to conventional random masking. Region-level attribution analysis reveals that brain volumes within the limbic region and cerebellum contribute most significantly to reconstruction fidelity and model representation. Our results demonstrate that masking anatomical regions during model pretraining not only enhances interpretability but also yields more robust and discriminative representations. In future work, we plan to extend this approach by evaluating it on additional neuroimaging datasets, and developing new loss functions explicitly derived from region-aware reconstruction objectives. These directions aim to further improve the robustness and interpretability of foundation models for functional neuroimaging.
\end{abstract}

\section{Introduction}
Hemodynamic signals, measured using techniques such as Functional Magnetic Resonance Imaging (fMRI) and functional Near-Infrared Spectroscopy (fNIRS), provide non-invasive markers of brain activity \cite{hart2012meta,pandey2024fnirsnet,adeli2025toward,adeliAb}. fMRI offers high spatial precision and has been widely used to investigate brain organization, functional connectivity, and cognitive processes. However, analyzing fMRI data remains challenging due to its high dimensionality, inter-subject variability, and heterogeneous acquisition protocols, which hinder reproducibility and generalization. The increasing availability of large and heterogeneous neuroimaging datasets has motivated the development of foundation models designed to learn generalizable representations across populations and tasks.

\quad Recent efforts in developing foundation models for fMRI have demonstrated the potential of large-scale pretraining to capture generalizable neural representations. Examples include Swin 4D fMRI Transformer (SwiFT) \cite{kim2023swiftswin4dfmri}, which introduces a transformer-based architecture for 4D fMRI data, and a graph transformer–based foundation model for functional connectivity networks \cite{WANG2026111988}, which integrates graph neural networks (GNNs) with transformer attention \cite{vaswani2023attentionneed} mechanisms to effectively capture connectivity patterns in brain networks. Within this paradigm, self-supervised learning has emerged as a particularly promising direction for harnessing large amounts of unlabeled data. Reconstruction-based approaches such as masked autoencoders (MAEs) \cite{MAE} mask parts of the input and train the model to recover them, thereby learning context-aware and transferable representations. Early demonstrations in other neuroimaging domains, such as EEG, have shown the utility of self-supervised approaches for seizure detection \cite{tang2022selfsupervisedgraphneuralnetworks}, with subsequent extensions \cite{das2022improvingselfsupervisedpretrainingmodels} incorporating random noise injection or single-channel removal to improve representation learning and robustness. In the fMRI domain, recent models such as BrainLM \cite{Caro2023.09.12.557460}, Brain-JEPA \cite{dong2024brainjepabraindynamicsfoundation}, and Self-supervised pre-training Tasks for fMRI time-series \cite{Zhou2024SelfSupervisedPT} adopt masking to capture spatiotemporal dynamics. While effective, these approaches typically apply masking after voxel signals are averaged into ROI-level time series through parcellation, sacrificing fine-grained spatial information and limiting localized reconstruction fidelity. In contrast, NeuroSTORM \cite{wang2025generalpurposefoundationmodelfmri} applies masking directly on full-resolution 4D fMRI volumes, but does not incorporate region-aware masking.

\quad In this paper, we propose a region-aware reconstruction strategy for fMRI foundation model pretraining. Our framework integrates anatomical regions directly in the masking process by using the Automated Anatomical Labeling atlas version 3 (AAL3) \cite{ROLLS2020116189} to selectively mask  brain regions within full-resolution 4D fMRI volumes. This preserves voxel-level spatial fidelity while encouraging the model to reconstruct localized and functionally meaningful signals. Evaluating on the ADHD-200 dataset \cite{adhd200}, we demonstrate that ROI-guided masking improves classification accuracy compared to conventional random masking. Furthermore, in line with prior ADHD literature implicating the limbic regions \cite{hart2012meta, VALERA2010359} and cerebellum \cite{article} in attentional control , we find that these regions play an important role in reconstruction fidelity.

\section{Methods}
\subsection{Self-supervised pretraining Framework}

Our self-supervised learning framework follows a two-stage process: a pretraining phase focused on masked voxel reconstruction, and a fine-tuning phase for prediction tasks. We adopt NeuroSTORM \cite{wang2025generalpurposefoundationmodelfmri} as the foundational model for fMRI, leveraging its encoder-decoder architecture for representation learning.

\quad In the pretraining stage, we introduce a region-of-interest (ROI) based masking strategy to improve the spatial specificity of self-supervised learning. Rather than applying random masking uniformly across space or time, we selectively mask spatiotemporal segments corresponding to anatomical or functional ROIs. This encourages the model to focus on reconstructing meaningful brain dynamics within targeted regions, thereby enhancing its ability to learn localized and functionally relevant features. The model is trained to reconstruct the masked segments of the input sequence without relying on supervision, enabling robust spatiotemporal representation learning. During the fine-tuning stage, we freeze the pretrained encoder weights and train a lightweight output head on top of the fixed representations, optimized for prediction tasks. By decoupling self-supervised representation learning from supervised classification and integrating ROI-guided masking during pretraining, our framework improves both generalization and interpretability in fMRI-based applications.
\begin{figure}[!htbp]
  \centering
  \includegraphics[width=1\linewidth]{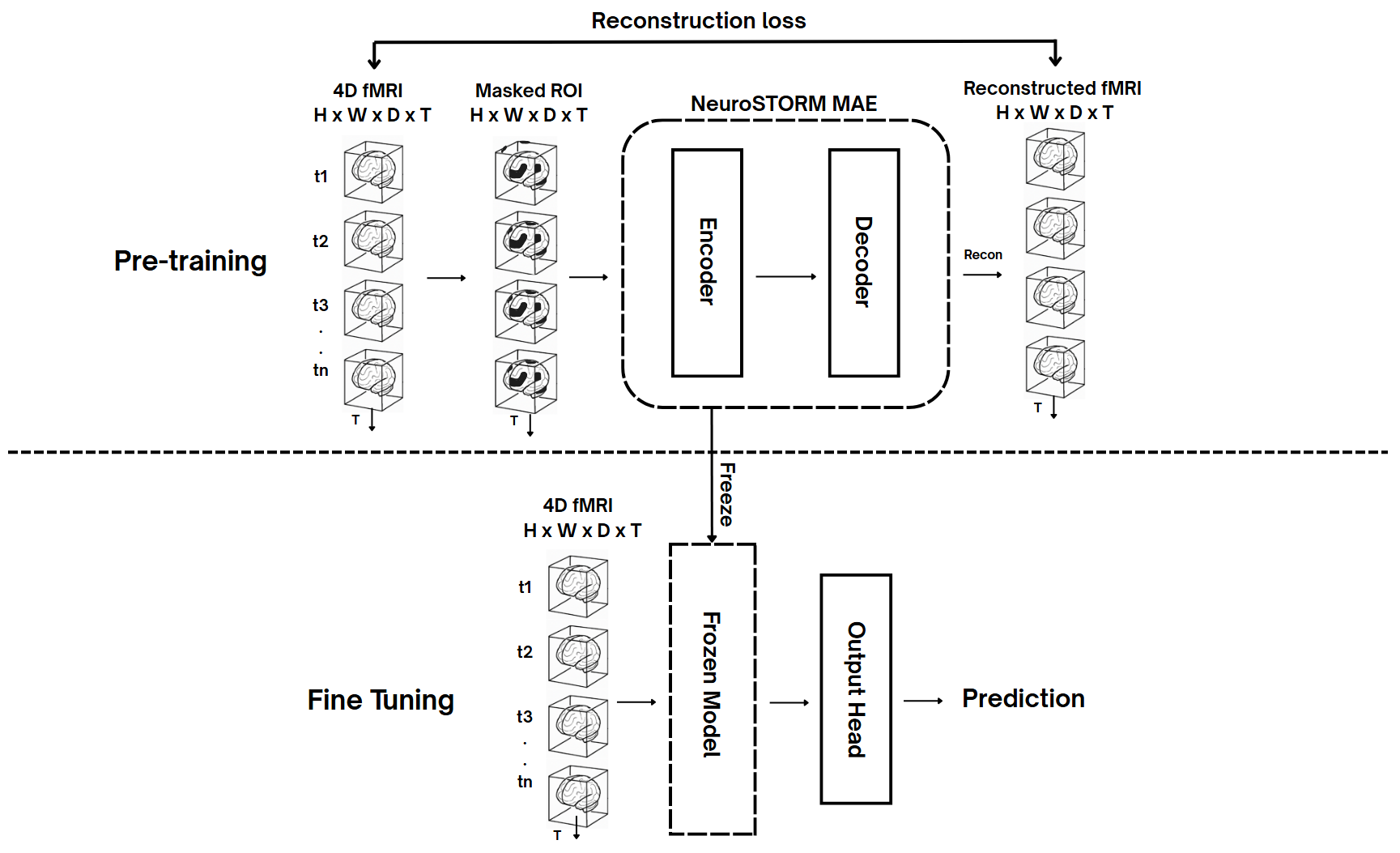}
  \caption{Illustration of ROI-guided masking strategy using AAL3 atlas. Specific anatomical regions are selectively masked during self-supervised pretraining.}
  \label{fig:roi-masking}
\end{figure}

\subsubsection{ROI-Based Masking}

Using the AAL3 atlas, we identify predefined anatomical regions and selectively mask a subset of these ROIs in the input 4D volume. This ROI-based masking extends the NeuroSTORM masking pipeline, which originally supports three masking strategies: random, random (randomly selected voxels in both space and time), random, tube (random spatial voxels masked consistently across all time points), and window, random (contiguous 3D spatial blocks with randomly masked time points). In our ROI-guided approach, the temporal masking dimension is equivalent to the tube setting, since we mask the same anatomical regions throughout the entire sequence.

\quad We train the model using targeted masking of regions within major anatomical domains, including the frontal, temporal, parietal, occipital lobes, cerebellum, limbic regions and subcortical structures. The proportion of brain volume masked varied substantially across ROI-based strategies, with regions defined according to the AAL3 atlas. Larger anatomical domains such as the frontal lobe (29.06\% of total brain voxels) and parietal lobe (16.65\%) masked a considerable portion of the signal, while smaller regions like the limbic regions (6.83\%) and Subcortical structures (3.86\%) involved proportionally less masking. In contrast, all of the existing NeuroSTORM masking strategies can include both brain and non-brain voxels, even though the overall proportion of the input masked is fixed at 10\%.

\begin{table}[!htbp]
\caption{Mask ratios for selected anatomical regions, defined according to the AAL3 atlas. The number of voxels masked and the corresponding percentage of total brain voxels are shown for each region.}
\label{tab:mask_ratios}
\centering

\begin{tabular}{lcc}
\toprule
\textbf{Mask} & \textbf{Number of voxels masked} & \textbf{Percentage of brain masked} \\
\midrule
    Frontal lobe      & 53,985  & 29.06\% \\
    Parietal lobe     & 30,927  & 16.65\% \\
    Temporal lobe    & 26,584  & 14.31\% \\
    Occipital lobe    & 26,358  & 14.19\% \\
    Cerebellum    & 24,414  & 13.14\% \\
    Limbic regions    & 12,690  & 6.83\%  \\
    Subcortical structures   & 7,165     & 3.86\% \\
\bottomrule
\end{tabular}
\end{table}

\subsection{Model Architecture: NeuroSTORM}

For fMRI analysis, we utilize NeuroSTORM (Neuroimaging Foundation Model with Spatial-Temporal Optimized Representation Modeling) \cite{wang2025generalpurposefoundationmodelfmri}, a state-of-the-art foundation model designed to learn directly from full 4D fMRI volumes. NeuroSTORM is pre-trained on over 28 million fMRI frames collected from more than 50,000 subjects, encompassing a broad demographic and clinical range across datasets such as UK Biobank \cite{UKB}, ABCD \cite{ABCD}, and HCP \cite{HCP}. This large-scale, multi-institutional corpus enables the model to generalize effectively across diverse downstream neuroimaging tasks. The model architecture is built upon a Shifted-Window Mamba (SWM) backbone, which integrates linear-time state-space modeling \cite{mamba} with shifted spatial window attention \cite{swin} to efficiently capture both local and global patterns in high-dimensional fMRI data. During self-supervised pretraining, NeuroSTORM adopts a masked autoencoding framework in which specific regions of the input are masked and reconstructed to facilitate robust representation learning.

\section{Dataset Description and Pre-processing}
\label{gen_inst}

We utilize the publicly available ADHD-200 dataset, which contains resting-state fMRI and phenotypic data from both typically developing controls and individuals diagnosed with Attention-Deficit/Hyperactivity Disorder (ADHD). The data is sourced from multiple sites and includes heterogeneous acquisition protocols, making it a valuable benchmark for evaluating the generalizability of models across diverse scanner conditions and subject populations. This dataset includes a total of 973 participants between the ages of 7 and 21 years.
 
\quad To ensure consistency across subjects and acquisition sites, we adopt a standardized fMRI pre-processing pipeline. All fMRI volumes are registered to the Montreal Neurological Institute (MNI152) template space, providing a common anatomical reference. Each volume is then resampled to an isotropic voxel resolution of 2 mm × 2 mm × 2 mm to enforce spatial consistency. To standardize input dimensions across samples, all 3D volumes are cropped or padded to a fixed spatial shape of 96 × 96 × 96. Temporal normalization is performed by resampling time series to a uniform repetition time (TR) of 0.8 seconds using linear interpolation. The resulting 4D volumes are then z-score normalized across non-background voxels and stored as per-frame tensors to support efficient loading in different tasks. For anatomical reference, the AAL3 atlas is also resampled and aligned to match the fMRI volumes.

\section{Experimental Settings}

We conducted self-supervised pretraining experiments on the publicly available ADHD200 dataset, comprising 973 subjects. The data was split into training, validation, and test sets using an 8:1:1 ratio. Pretraining was performed for 20 epochs with a batch size of 24, using a model with approximately 7.7 million parameters. The optimization was performed using the AdamW optimizer with a learning rate of 5e-5. Reconstruction loss was computed using mean squared error (MSE) between the masked and predicted fMRI volumes, and all experiments were run on a high-performance computing cluster equipped with 3*NVIDIA A100-SXM4-80GB GPUs. During training, each GPU utilized up to 60 GB of memory, with utilization reaching 70–80\%, indicating efficient use of computational resources. On this setup, pretraining required approximately 16 hours to complete, while fine-tuning took approximately 2 hours to run.

\section{Results}
\label{headings}

 During pretraining, we observed distinct differences in mean squared error (MSE) across different masking strategies. Random tube masking yielded the lowest MSE (0.0311), significantly lower than ROI-based masking approaches such as Cerebellum (0.0601), Frontal lobe (0.0759), Temporal lobe (0.0699), and Parietal lobe (0.0860). This discrepancy can be attributed to the nature of the random masking strategy, which often includes non-brain regions (e.g., background voxels) in the masked input. These non-brain regions typically exhibit low or zero signal and are thus easier to reconstruct, leading to artificially low reconstruction errors. In contrast, ROI-based masking restricts the masked areas to anatomically defined brain regions, ensuring that the model focuses solely on reconstructing meaningful neural signals.

\begin{table}[!htbp]
\caption{Reconstruction loss and ADHD classification performance (Accuracy and AUCROC) for distinguishing healthy controls from individuals diagnosed with ADHD under existing masking strategies.}
\label{tab:baseline_results}
\centering
\begin{tabular}{c c ccc}
\toprule
\textbf{Spatial Mask} & \textbf{Time Mask} & \textbf{Reconstruction loss} & \textbf{ACC} & \textbf{AUCROC} \\
\midrule
Random   & Random   & 0.0263 & 62.76\% & 0.678 \\
Window   & Random & 0.0322 & 62.76\% & 0.651 \\
Random   & Tube   & \textbf{0.0311} & \textbf{63.82\%} & \textbf{0.680} \\
\bottomrule
\end{tabular}
\end{table}

\begin{table}[!htbp]
\caption{Reconstruction loss and ADHD classification performance (ACC and AUCROC) under atlas-based masking strategies.
Each row corresponds to masking all voxels within the specified AAL3 anatomical region. 
For all atlas-based experiments, the time mask is fixed to \textit{tube}. *Note that MSE values across different ROIs are not directly comparable, since the number of voxels masked varies by region. Larger regions involve masking more voxels, which leads to higher reconstruction loss, whereas smaller regions mask fewer voxels and yield lower losses. The last three rows correspond to partial regional masking (50\% of voxels within the specified region).
}

\label{tab:atlas_results}
\centering
\begin{tabular}{lccc}
\toprule
\textbf{Atlas Region Masked} & \textbf{Reconstruction loss*} & \textbf{ACC} & \textbf{AUCROC} \\
\midrule
Frontal lobe        & 0.0759 & 65.95\% & 0.671 \\
Parietal lobe       & 0.0860 & 64.89\% & 0.716 \\
Temporal lobe       & 0.0699 & 64.84\% & 0.691 \\
Occipital lobe      & 0.0893 & 64.89\% & 0.658 \\
Cerebellum          & 0.0601 & 68.05\% & 0.720 \\
Limbic regions      & \textbf{0.0455} & \textbf{68.08\%} & \textbf{0.752} \\
Subcortical structures & 0.0482 & 67.02\% & 0.685 \\
\midrule
Frontal lobe + Cerebellum &  0.0683 & 65.95\% & 0.711 \\
Cerebellum + Limbic regions & 0.0552 & 68.08\% & 0.711 \\
Limbic regions + Frontal lobe & 0.0742 & 65.42\% & 0.698 \\
\midrule
Frontal lobe (50\%) & 0.0570 & 65.95\% & 0.701 \\
Limbic regions (50\%) & 0.0694 & 67.02\% & 0.721 \\
Limbic regions (50\%) + Cerebellum (50\%) & 0.0585 & 67.02\% & 0.699 \\
\bottomrule
\end{tabular}
\end{table}

\quad The ROI-based pretraining yielded superior classification accuracy (ACC) and AUCROC compared to random masking. Notably, cerebellum and limbic regions masking substantially improved both ACC (68.05\%, 68.08\% respectively) and AUCROC (0.720, 0.704 respectively) relative to random tube masking (63.82\% ACC, 0.680 AUCROC). These results indicate that masking anatomical regions during pretraining encourages the model to learn more discriminative and functionally relevant representations, even at the expense of higher reconstruction error. In addition to single-region masking, we also evaluated binary region combinations (frontal lobe + cerebellum, limbic regions + cerebellum, limbic regions + frontal lobe), whose results are summarized in Table~\ref{tab:atlas_results}. To further examine the effect of partial region removal, we experimented with masking 50\% of specific regions. As shown in Table~\ref{tab:atlas_results}, masking half of the voxels in the frontal and limbic regions, either independently or jointly with the cerebellum, did not yield consistent improvements in classification accuracy or AUCROC. This suggests that partial regional masking alone does not fundamentally alter performance. Importantly, these findings indicate that unmasked regions may differentially contribute to model prediction, motivating the need for systematic investigations into how regional dependencies influence reconstruction and classification accuracy. Random masking by itself is not sufficient; instead, carefully designed masking strategies are required. 

\section{Conclusion, Limitations and Future Directions}
In this work, we introduced a region-aware reconstruction strategy for fMRI foundation model pretraining by extending the NeuroSTORM framework with ROI-guided masking based on the AAL3 atlas. Our ROI-guided masking strategy achieved a 4.23\% improvement in classification accuracy compared to conventional random masking, with single-region experiments showing that cerebellum and limbic regions were particularly effective. These results demonstrate that incorporating anatomical structure into the masking process enhances the model performance.\\ \\
Our study is limited to the ADHD-200 dataset, and in future work we plan to evaluate the approach on larger and more diverse datasets to more rigorously assess generalizability. We also aim to design region-aware loss functions that explicitly incorporate anatomical structure, moving beyond generic reconstruction objectives. In addition, we plan to investigate the impact of ROI-guided masking across a wider range of tasks and explore adaptive, data-driven masking strategies that dynamically select informative regions during pretraining.











\bibliographystyle{unsrtnat}   
\bibliography{references}      

\end{document}